# AUTOMATIC SEGMENTATION OF MANIPURI (MEITEILON) WORD INTO SYLLABIC UNITS

Kishorjit Nongmeikapam[1], Vidya Raj RK[2], Oinam Imocha Singh[3] and Sivaji Bandyopadhyay[4]

[1]Department of Computer Science & Engineering, Manipur Institute of Technology, Manipur University, Imphal, India
`kishorjit.nongmeikapa@gmail.com`
[2,3] Department of Computer Science, Manipur University, Imphal, India
`vidyarajrk@yahoo.com, imocha2007@rediffmail.com`
[4]Department of Computer Science & Engineering, Jadavpur Univeristy, Jadavpur, Kolkata, India
`sivaji_cse_ju@yahoo.com`

*ABSTRACT*

*The work of automatic segmentation of a Manipuri language (or Meiteilon) word into syllabic units is demonstrated in this paper. This language is a scheduled Indian language of Tibeto-Burman origin, which is also a very highly agglutinative language. This language usages two script: a Bengali script and Meitei Mayek (Script). The present work is based on the second script. An algorithm is designed so as to identify mainly the syllables of Manipuri origin word. The result of the algorithm shows a Recall of 74.77, Precision of 91.21 and F-Score of 82.18 which is a reasonable score with the first attempt of such kind for this language.*

## 1. INTRODUCTION

A **syllable** is a basic unit of written and spoken language. It is a unit consisting of uninterrupted sound that can be used to make up words. For example, the word *unladylike* has four syllables: *un, la, dy* and *like*. These will be marked here as *un/la/dy/like*. The syllable is a structural unit and within that structure we can identify a sequence of consonants (C) and vowels (V) or on basis of onset (beginning of a syllable, either a consonant or a semivowel), peak (nucleus of the syllable, vowels) and coda (sound which comes after the peak, generally consonants).

So far works of word segmentation to syllabic units for Manipuri Language is not reported and this is the first work of such kind upto the authors' knowledge. This Manipuri Language or simply Manipuri is a highly agglutinative Schedule Indian Language. This language usage two scripts one is a Bengali Script and another is the original Manipuri script that is the Meitei Mayek (Script).

In this work an algorithm is being designed in order to identify the syllabic units automatically. The algorithm is suitable mainly for the Meitei Mayek. It is because it has less character compared with the Bengali Script and easy to distinguish the original and loan words formation with the syllabic units.

Different works related to syllable or syllabic units can be found. Phoneme monitoring, syllable monitoring and lexical access are mention in [1] and a

129

comparison on syllabic and segmental perception is reported in [2]. In [3] the syllable analysis is done to build a dictation system in Telugu language. An implementation of prosodic unit or the pseudo-syllable in speech recognition is reported in [4] and also rhythmic unit extraction and modelling for automatic language identification is reported in [5]. An implementation of word and syllable models for German text-to-speech synthesis is found in [6]. In speech recognition syllabic unit is used for speech recognition, this is found in [7].

Work on the role of strong syllables in segmentation for lexical access is reported in [8]. In multiple language like english, german, french and spanish work of syllabic features and phonic impression is reported in [9].

The paper is organized with motivations and related works are discussed in Section 2 which is followed by the discussion of Meitei Mayek or the Meiteilon Alphabets in Section 3, Section 4 is about the system design and the algorithms of the syllabic units, Section 5 make a brief discussion about the experiment and the evaluation, Section 6 draws the conclusion and the future works road map.

## 2. MOTIVATION

The necessity of designing an efficient morphological analyser of this language very much motivates this work. The design of a morphological analyser is still a complex task for this language since it's a very highly agglutinative language. The believed of some relationship with the identification of syllabic unit and the morpheme structure in Manipuri also motivates this work.

The use of syllabic units could be useful in the text to speech conversion or in other speech conversion works. These are mention in much of the published works like in [4]-[7]. So this is the other factor of motivation. This work could also be helpful in development of the lexicon resources.

The design of spell checker in Manipuri may definitely require such work in future since no efficient spell checker is design till date for this language. This is because this language is a morpheme reach language.

The works of [10] is design as a light weight stemmer for Manipuri but the central idea shows about the segmentation of the affixes so that the root words can be identified. This work motivates the possibility of segmenting the words into syllabic units.

## 3. MEITEI MAYEK OR THE MEITEILON ALPHABETS

The Manipuri has its original script but there was an era where the language is being influence by the Bengali Script. The revival of the original script bring another controversy in the number of characters in the scripts of this language thus no much advancement could have been done in the computational research works. Our work is based on the 27 scripts approved by the State Govt. which is also published in [12].

Before we go deeper into other things it is very important to get familiar with the characters that constitute the script. Like other language the characters used can be group into vowels, consonants, numerical figures and other symbols. The characters used in Meiteilon (Manipuri language) can be classified into five different categories.

- Iyek Ipee : *(See Table 1)* This character set consists of 27 letters which are mainly major consonants, out of which three are used to produce vowel sounds (ꯁ, ꯍ, ꯑ). This category is considered as major consonants in the sense that letters are used in their full form at the initial position of a syllabic unit. Moreover, associations with Cheitap Iyek are permitted with these characters only.

- Cheitap Iyek (Matras): (*See Table 2*) These are associative symbols which can be found only in association with Iyek Ipee character sets. Association with Iyek Ipee characters follow a one to one relationship i.e. no two (or more) symbols is found to be associated with one letter in Iyek Ipee. Consecutive occurrence is also not permitted.
- Cheising Iyek (Numerals): (*See Table 3*) This set contains the numeral figures and follow the decimal system.
- Lonsum Iyek: (*See Table 4*) There are 8 characters in this set and these characters can be considered to be derivative form of 8 distinct letters in Iyek Ipee. In one sense, these letters can be regarded as half consonants as they cannot be associated with any symbols in Cheitap Iyek and cannot initiate formation of a syllable. This character set can only be present in the syllabic final position. Recurrence or clusters of these characters i.e. consecutive occurrence of these characters are also not permitted in the language.
- Khudam Iyek (Symbols): Usage of special characters is limited in this language and as such few symbols suffice the need in expression.
  Examples:

  '‖'   - Cheikhei (Full Stop)

  '.'   - Lum Iyek (Sign of intonation) eg. ꯆꯥ.ꯕ (*cha.ba* (to eat)) falling intonation and ꯆꯥꯕ (*cha.ba* (swimming/floating)) rising intonation.

  '_'   - Apun Iyek (Sign of Ligature) eg. ꯆꯝꯁꯪ (*cham.pra* (lemon))

Other symbols are as internationally accepted symbols.

Table 1.  Iyek Ipee characters in Meitei Mayek.

| **Iyek Ipee** | | | |
|---|---|---|---|
| ꯀ (kok) | ꯁ (Sam) | ꯂ (Lai) | ꯃ (Mit) |
| ꯄ (Pa) | ꯅ (Na) | ꯆ (Chil) | ꯇ (Til) |
| ꯈ (Khou) | ꯉ (Ngou) | ꯊ (Thou) | ꯋ (Wai) |
| ꯌ (Yang) | ꯍ (Huk) | ꯎ (Un) | ꯏ (Ee) |
| ꯐ (Pham) | ꯑ (Atia) | ꯒ (Gok) | ꯓ (Jham) |
| ꯔ (Rai) | ꯕ (Ba) | ꯖ (Jil) | ꯗ (Dil) |
| ꯘ (Ghou) | ꯙ (Dhou) | ꯚ (Bham) | |

Table 2.  Cheitap Iyek of Meitei Mayek.

| **Cheitap Iyek** | | | |
|---|---|---|---|
| ꯣ (ot nap) | ꯤ (inap) | ꯥ (aatap) | ꯦ (yetnap) |
| ꯧ (sounap) | ꯨ (unap) | ꯩ (cheinap) | ꯪ (nung) |

Table 3. Cheising Iyek or numerical figures of Meitei Mayek

| Cheising Iyek(Numeral figure) | | | |
|---|---|---|---|
| ꯱(ama) | ꯲(ani) | ꯳(ahum) | ꯴(mari) |
| ꯵(manga) | ꯶(taruk) | ꯷(taret) | ꯸(nipal) |
| ꯹(mapal) | ꯱꯰(tara) | | |

Table 4. Lonsum Iyek of Meitei Mayek

| Lonsum Iyek | | | |
|---|---|---|---|
| ꯛ (kok lonsum) | ꯜ (lai lonsum) | ꯝ (mit lonsum) | ꯞ (pa lonsum) |
| ꯟ (na lonsum) | ꯠ (til lonsum) | ꯡ (ngou lonsum) | ꯢ (ee lonsum) |

## 4. SYSTEM DESIGN AND THE ALGORITHM FOR SYLLABIC UNITS IDENTIFICATION

Keeping in mind about the patterns of the syllabic units in Manipuri words the system is design with a flowchart and algorithms which are discuss below:

Figure 1. The Flowchart of the word segmentation into the syllabic unit

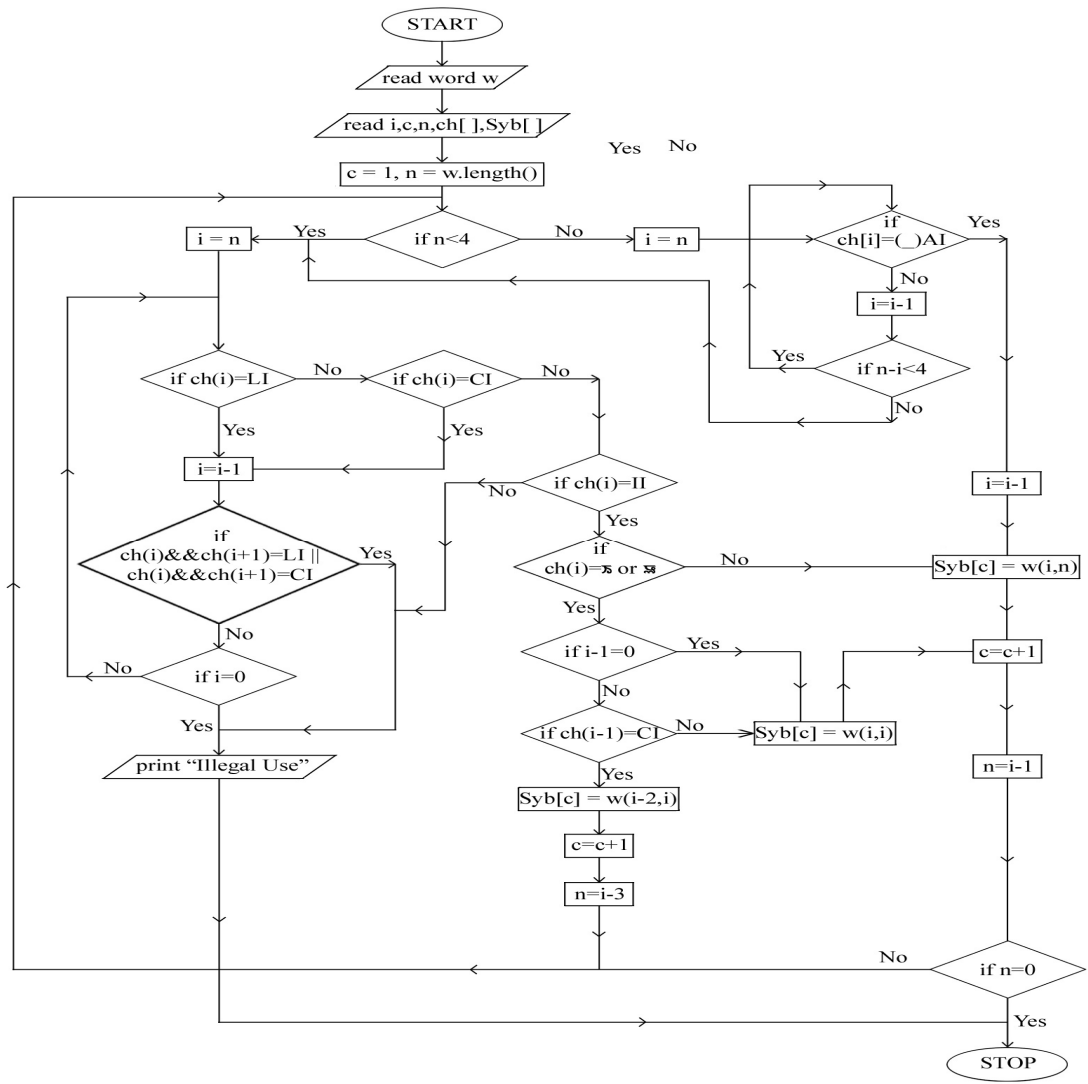

Symbol used in the Flowchart:
**w.length()**= Length of the word w
**II** = Ipee Iyek
**LI** = Lonsum Iyek
**CI** = Cheitap Iyek
**AI** = Apun Iyek (_)
**w(i,j)** = sub string extracted from index i to j

Algorithm 1 provides the base foundation in which the Input File is read line by line and every line is tokenized and every tokens or word is provided for syllable extraction. Then from the stack, where the syllables for every word are stored, the unit/mono syllables are again written into the Output File.

1. **AGORITHM 1:** SEGMENT(Inputfile)
2.     str ← *readline*(Inputfile)
3.     while str != null
4.         arr[ ] ← *takenize*(str)
5.         i ← 0
6.         word ← arr[i]
7.         while word != null
8.             *sybextract*(word)
9.             i++
10.            syb ← *pop*()
11.            while syb != null
12.         *write*(syb)
13.         end of while
14.         word ← arr[i]
15.         end of while
16.         str ← *readline*(Inputfile)
17.     end of while

Algorithm 2, when called by Algorithm 1, takes a string parameter (word) and segments the word into unit syllables. Segmentation is done depending on the script based rules and the syllabic structures defined in this paper. For every syllable extracted it is pushed down into a stack object defined for the particular word. The extraction process starts from the left most syllables, thus suitable for storing in a stack. The following gives the details about the second algorithm which is used in this system.

1. **ALGORITHM 2:** SYBEXTRACT(word)
2.     len ← word.length
3.     n ← len-1
4.     i ← n
5.     ch[ ] ← word
6.     while n!=-1
7.       if n>=3
8.         i ← n
9.         flag ← true
10.        while *cmpr*(ch[i],AI) = false
11.          i--

```
12.            if n-i < 4
13.               continue
14.            end of if
15.            else
16.               flag ← false
17.               break
18.            end of else
19.         end of while
20.         if flag = true
21.            i--
22.            push(word.substring(i,n+1))
23.            n ← i-1
24.            if n=-1
25.               return
26.            end of if
27.            else
28.               continue
29.            end of else
30.         end of if
31.      i ← n
32.      while true
33.         flag2 ← false
34.         if cmpr(ch[i],LI) = true
35.            flag2 ← true
36.         end of if
37.         else
38.            if cmpr(ch[i],CI) = true
39.               flag2 ← true
40.            end of if
41.            else
42.               if cmpr(ch[i],II) = true
43.                  if cmpr(ch[i],SC) = true
44.                     if i-1 = -1
45.                        push(word.substring(i,i+1))
46.                        n ← i-1
47.                        if n = -1
48.                           return
49.                        end of if
50.                        else
51.                           break
52.                        end of else
53.                     end of if
54.                     else
55.                        if cmpr(ch[i-1],CI) = true
56.                           push(word.substring(i-2, i+1))
57.                           n ← i-3
58.                           break
59.                        end of if
60.                        else
61.                           push(word.substring(i,i+1))
62.                           n ← i-1
63.                           if n = -1
64.                              return
```

134

```
65.                    end of if
66.                  else
67.                      break;
68.                  end of else
69.                end of else
70.              end of else
71.            else
72.              push(word.substring(i, n+1))
73.              n ← i-1
74.              if n = -1
75.                return
76.              end of if
77.              else
78.                  break;
79.              end of else
80.            end of else
81.          end of if
82.          else
83.            push(word)
84.            return
85.          end of else
86.        end of else
87.      end of else
88.        if flag2 = true
89.          i--
90.          if((cmpr(ch[i],LI)) and (cmpr(ch[i+1],LI)) or ((cmpr(ch[i],CI)) and (cmpr(ch[i+1],CI)) = true
91.              push(word)
92.              return
93.        end of if
94.          else
95.            if i = -1
96.              push(word)
97.              return
98.            end of if
99.            else
100.               continue
101.            end of else
102.        end of else
103.      end of if
104.   end of while
105. end of while
```

## 5. EXPERIMENT AND EVALUATION

The experiment is conducted with a gold standard corpus of 6000 wordforms. The corpus is clean with a linguistic knowledge so that a better output is yield. The system is made to compare with the linguistic syllabic patterns and the computational syllabic patterns output. The following sub sections discusses about the experimental result followed by the discussion with linguistic patterns, the computational output patterns and the comparison between both the patterns.

### 5.1. Experimental result

The result is measured with the parameter of Recall, Precision and F-score. The definitions of the terms are defined as follows:

Recall,

$$R = \frac{No\ of\ correct\ syllables\ given\ by\ the\ system}{No\ of\ correct\ syllables\ in\ the\ text}$$

Precision,

$$P = \frac{No\ of\ correct\ syllables\ given\ by\ the\ system}{No\ of\ syllables\ given\ by\ the\ system}$$

F-score,

$$F = \frac{(\beta^2 + 1)\ PR}{\beta^2 P + R}$$

Where $\beta$ is one, precision and recall are given equal weight.

In this work the system shows a Recall of **74.77**, Precision of **91.21** and F-Score of **82.18**. The analysis of the output can be discussed with the comparison of the linguistic patterns and the computational syllabic pattern outputs.

### 5.2. Manipuri Linguistic Syllabic Pattern

The Syllables in Manipuri can be divided into three parts; onset (beginning of a syllable, either a consonant or a semivowel), peak (nucleus of the syllable, vowels) and coda (sound which comes after the peak, generally consonants). In every syllable there must be a peak.

In Manipuri there may not be an onset or coda in the syllabic system. Referring to section 2.4.1 of [1], the syllabic structure, the author has stated that the syllables can be of six forms which are listed as Classification A below:

**Classification A**

1) V
2) VC
3) CV         where,
4) CCV        V = Syllabic peaks, vowels
5) CVC        C = Syllabic margins, consonants
6) CCVC

### 5.3. Computational Syllabic Patterns

After running the system, some observations can be drawn from the computational output. In the algorithm adopted by this system the syllables are segmented based on characters and thus the patterns observed also are on the basis of the characters. The observation shows 11(eleven) syllabic patterns and the patterns found are as follows:

**Classification B**

1) V
2) CV
3) C
4) VVV
5) CVC
6) CC
7) CVV

8) VV
9) VVC characters
10) VC characters
11) CCVC

where,
V = vowel
C = consonant

## 5.4. Comparative study of both the patterns

The Linguistic approach shows syllabic patterns of six and the computational output shows a total of eleven patterns. It may seems the later classification shows more variety of patterns, but after minute cross verification it is not so. Either the classifications are one or otherwise the same, the only difference lies in the interpretation. To elaborate this discussion let us consider the characters that are considered as consonants and vowel characters.

**Consonant Characters**

The Iyek Ipee characters with the exception of 'ꯑ', 'ꯏ', and 'ꯎ' along with all the Lonsum Iyek characters are considered as consonant characters.

ꯀ ꯁ ꯂ ꯃ ꯄ ꯅ ꯆ ꯇ ꯈ ꯉ ꯊ ꯋ ꯌ ꯍ ꯐ ꯑ ꯒ ꯓ ꯔ ꯕ ꯖ ꯗ ꯘ ꯙ ꯚ ꯛ (Iyek Ipee characters without 'ꯑ', 'ꯏ', and 'ꯎ')

ꯛ ꯜ ꯝ ꯞ ꯟ ꯠ (Lonsum Iyek characters)

**Vowel Characters**

The Cheitap Iyek characters along with 'ꯑ', 'ꯏ', and 'ꯎ' are considered as vowels. The three of the Iyek Ipee characters 'ꯑ', 'ꯏ', and 'ꯎ' which are excluded in the previous consonant characters list are considered to be the vowel characters.

ꯣ, ꯤ, ꯥ, ꯦ, ꯧ, ꯨ, ꯩ and ꯪ (matras) and ꯑ ꯏ ꯎ (three of the Iyek Ipee characters ꯑ, ꯏ, and ꯎ).

## 5.5. Pattern Description and Ambiguity Conditions

Linguistically, the class of syllables 'V' has a peak but onset and coda are absent. For example; 'ꯎ'(u) meaning 'tree'. Here there is no ambiguity and since only one character is present, this category of syllables is grouped as Classification B.1.

In case of syllables like 'ꯑꯣ'(o) which is an exclamation word, it linguistically consists of the peak only but character wise there are two distinct characters, and hence in Classification A it will be classified as V and in Classification B, as VV. In the Classification B the patterns are sometimes observed as VVV for VC and CVV for CVC, it is because the last V in VVV or V in CVV are observed to be semi vowel. Table 5 show the comparative study of Classification A and Classification B with ambiguities and examples.

Table 5. Comparison of the patterns and examples

| CLASS A | CLASS B | EXAMPLES | SOURCE WORD | MEANING |
|---|---|---|---|---|
| V | V | ꯁ | ꯁ | blood |
|   | VV | ꯑꯣ | ꯑꯣ | exclamation |
| VC | VC | ꯏꯁ | ꯏꯁ | ass |
|   | VVC | ꯑꯣꯁ | ꯑꯣꯁꯕ | harass |
|   | VVV | ꯑꯣꯁ | ꯑꯣꯁꯨ | owning |
| CV | CV | ꯃꯤ | ꯃꯤ | eye brows |
|   | C | ꯕ | ꯆꯕ | eat |
| CCV | -- | ꯀꯋꯥ | ꯀꯋꯥ | beetle nut |
| CVC | CVC | ꯄꯥꯠ | ꯄꯥꯠ | lake |
|   | CVV | ꯄꯥꯏ | ꯄꯥꯏ | fly |
|   | CC | ꯆꯠ | ꯆꯠꯂꯤ | going |
| CCVC | CCVC | ꯀꯋꯥꯛ | ꯀꯋꯥꯛ | crow |

The other two main reasons behind these ambiguities are as follows:
1) In Meiteilon (Manipuri language) some vowel sound comprise of more than one character, for example; o = ꯑꯣ. These can be observed in the table 5 Class B where the patterns are denoted as VV instead of V.
2) In some syllables like 'ꯄ', which is basically a suffix, a single character is used to denote it but linguistically it contains a hidden 'ə' (schwa) making the syllable as 'Pə'.

## 7. CONCLUSION

The segmentation of syllabic unit will bring very close to the morphological analyser of Manipuri language, which is so far a tough job. Another future implementation could be in the spell checker of Manipuri, which is not yet tried. This morphologically rich language can used this algorithm for other lexical resource development work. The nature of agglutinative makes the task tougher. Apart from it may be a better approach in future for the implementation of text to speech conversion.
The result shows a Recall of **74.77%**, Precision of **91.21%** and F-Score of **82.18%** which is a reasonably a good score with the first attempt of such kind for this language. More works could be done for the improvement of the score and can think of other implementations for such works.

**Authors**

Kishorjit Nongmeikapam is working as Asst. Professor at Department of Computer Science and Engineering, MIT, Manipur University, India. He has completed his BE from PSG college of Tech., Coimbatore and has completed his ME from Jadavpur University, Kolkata, India. He is presently doing research in the area of Multiword Expression and its applications. He has so far published 22 papers and presently handling a Transliteration project funded by DST, Govt. of Manipur, India.

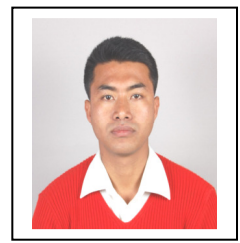

Vidya Raj R K is presently a student of Manipur University. He is pursuing his MCA in Dept. of Computer Science. His area of interest is NLP.

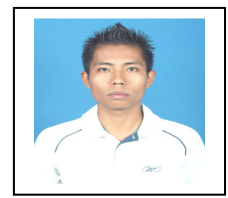

Oinam Imocha Singh is the Associate Professor and HOD and in Department of Computer Science, Manipur University, Imphal, Manipur, India. Area of interest include Artificial Intelligence, NLP, Image Processing, Machine Learning, Mathematical Modelling etc. Published 5 international papers in the fields of Image Processing and Theorem Proving.

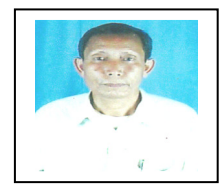

Professor Sivaji Bandyopadhyay is working as a Professor since 2001 in the Computer Science and Engineering Department at Jadavpur University, Kolkata, India. His research interests include machine translation, sentiment analysis, textual entailment, question answering systems and information retrieval among others. He is currently supervising six national and international level projects in various areas of language technology. He has published a large number of journal and conference publications.

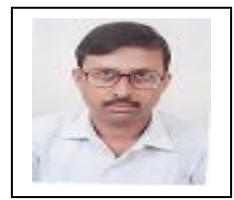